\documentclass{article}

\usepackage{arxiv}

\usepackage[utf8]{inputenc} 
\usepackage[T1]{fontenc}    
\usepackage{url}            
\usepackage{booktabs}       
\usepackage{amsfonts}       
\usepackage{nicefrac}       
\usepackage{microtype}      
\usepackage{lipsum}
\usepackage{amsmath}
\usepackage{multirow}
\usepackage{graphicx}
\graphicspath{ {./images/} }

\title{Interpretation of Deep Learning Model in Embryo Selection for In Vitro Fertilization (IVF) Treatment}

\author{
 Kodali Radha \\
  Division of Pediatric Neurology\\ Department of Pediatrics\\ University of Tennessee Health Science Center\\ Memphis, 38163, TN, USA\\
  \texttt{rkodali@uthsc.edu} \\
   \And
Dhulipalla Venkata Rao\\\
 Department of Electronics \& Communication Engineering \\Velagapudi Ramakrishna Siddhartha Engineering College\\ Kanuru, Vijayawada, 520007 , Andhra Pradesh, India \\
  \And
  Tatavarty Venkata Siva Kishor\\\
 Department of Enterprise Data and Analytics\\ Converse Inc., Boston \\MA, 02180, USA \\
  \And
  Nadakuditi Madhavi\\\
 Department of Electronics \& Communication Engineering \\Velagapudi Ramakrishna Siddhartha Engineering College\\ Kanuru, Vijayawada, 520007 , Andhra Pradesh, India \\
  \And
   Thiruveedhula Bharadwaj\\\
 Department of Electronics \& Communication Engineering \\Velagapudi Ramakrishna Siddhartha Engineering College\\ Kanuru, Vijayawada, 520007 , Andhra Pradesh, India \\
  \And
  Suryanarayana Gunnam\\\
 Department of Electronics \& Communication Engineering \\Velagapudi Ramakrishna Siddhartha Engineering College\\ Kanuru, Vijayawada, 520007 , Andhra Pradesh, India \\
  \And
 Durga Prasad Bavirisetti \\
Department of Computer Science\\ Norwegian University of Science and Technology\\ 7034 Trondheim, Norway
  \texttt{durga.bavirisetti@ntnu.no} \\
 \And
  Gogulamudi Pradeep Reddy \\
  Department of Information and Communication Technology\\ Manipal Institute of Technology\\ Manipal Academy of Higher Education,\\Manipal, Karnataka 576104, India.
}

\begin{document}
\maketitle
\begin{abstract}
Infertility has a considerable impact on individuals' quality of life, affecting them socially and psychologically, with projections indicating a rise  in the upcoming years. In vitro fertilization (IVF) emerges as one of the primary techniques within economically developed nations, employed to address the rising problem of low fertility. Expert embryologists conventionally grade embryos by reviewing blastocyst images to select the most optimal for transfer, yet this process is time-consuming and lacks efficiency. Blastocyst images provide a valuable resource for assessing embryo viability. In this study, we introduce an explainable artificial intelligence (XAI) framework for classifying embryos, employing a fusion of convolutional neural network (CNN) and long short-term memory (LSTM) architecture, referred to as CNN-LSTM. Utilizing deep learning, our model achieves high accuracy in embryo classification while maintaining interpretability through XAI. To enhance the robustness of blastocyst images, data augmentation techniques are utilized, while feature extraction is facilitated through CNN-LSTM architecture. The model is designed as a binary classifier, distinguishing between two classes: good and poor embryos. The XAI approach employed is local interpretable model-agnostic explanations (LIME), enabling the visualization of the model’s decision-making process. Notably, our proposed CNN-LSTM model achieves a notable accuracy of 90\% before augmentation and an impressive 97.7\% after augmentation, highlighting its potential for clinical application. Incorporating XAI ensures clear and interpretable insights into the model’s decision-making process, enhancing its utility in real-time clinical settings.
\end{abstract}


\maketitle

\section{Introduction}
Infertility is a frequent reproductive health issue impacting millions globally, resulting in psychological, physical, social and economic challenges for individuals attempting to conceive. The world health organization (WHO) estimates that globally, approximately 48 million couples and 186 million individuals struggle with infertility \cite{mohamed2023automated}. Infertility significantly affects the individual's quality of life, both socially and psychologically, and is estimated to expand in the coming years. According to WHO, one in every four couples experiences infertility problem \cite{vaidya2021time}.  To tackle this problem, assisted reproductive technology (ART) were formulated, providing assistance to individuals and couples in overcoming difficulties in conception. The utilization of ART among infertile couples has been rising steadily by 5\% to 10\% annually \cite{ravitsky2019forgotten}.

In vitro fertilization (IVF) emerges as a prominent method of ART, adopted to address various fertility complications and enhance opportunities for future parenthood in situations with fertility problems. IVF, a fertility intervention, where a mature egg (oocyte) and sperm are united outside the body, particularly within a dedicated laboratory environment. An IVF cycle comprises a series of consecutive steps, encompassing gamete selection, embryo cultivation, quality assessment, and implantation, all with the ultimate goal of achieving pregnancy \cite{tran2020swot}.
The fertilized egg, referred to as the embryo, is cultured in a controlled environment for up to seven days \cite{niu2020day} before being transferred to the woman's uterus, enhancing the chances of a successful pregnancy. Currently, the main techniques used for IVF fertilization are conventional morphological assessment (CMA) and embryonic morpho-kinetics (EMK) using time-lapse (TL) platforms \cite{martinez2017inter}. Both approaches incorporate evaluating and selecting embryos, with TL utilizing  incubators equipped with integrated microscopes programmed to capture images automatically at predefined intervals and magnification levels, whereas CMA relies on visual monitoring daily. However, different selection methods may emerge from various IVF procedures, some of which could raise concerns regarding their quality, particularly during the developmental blastocyst stage. The fertilized egg evolves into a human embryo (zygote) and goes through four stages: a pronucleate oocyte (PO), cleavage stage (CS), morula (M), and blastocyst (BC) \cite{isa2023image}, considering the hours following insemination and the envisaged characteristics at each developmental stage. In the IVF process, the quality of fertilized eggs is crucial, as selecting a high-quality fertilized egg could significantly enhance the chances of a successful pregnancy rate \cite{mastenbroek2011embryo} prior to the embryo's transfer into the uterus. A fertilized PO, after day 5 following insemination, is recognized as a BC. A blastocyst is characterized by the presence of a fluid-filled cavity, known as the blastocoel, located at the embryo's center. Surrounding this cavity is a single layer of cells called the trophectoderm, which is essential in the formation of numerous extraembryonic tissues. Blastocysts exhibit greater potential for implantation compared to embryos at the cleavage stage (embryonic day 3) \cite{tandulwadkar2019optimising}. Research has indicated that continuing embryo culture up to day 5 results in a higher chance of successful delivery \cite{glujovsky2022cleavage}. Hence, evaluating the embryo on day 5 (blastocyst) is significant. Embryo grading on day 5 follows gardner system, where the grade is decided by evaluating the characteristics  of both inner cell mass (ICM), and the trophectoderm (TE). Key factors influencing the blastocyst quality include the rate of embryo size expansion, and the analysis of both ICM and TE cells. 

Embryologists primarily approve that the differentiation in morphological structure between the TE and ICM serves as a key indicator of the developmental advancement of the formed blastocyst. The conventional method for evaluating embryo quality is notably time-consuming and relies on visual evaluation of embryo morphology by embryologists through a microscope, making it highly practice-dependent and subjective. Multiple studies have aimed to create automated grading systems, implementing methods such as convolutional neural networks to overcome the limitations associated with manual grading of embryos. However, they are not suitable for the real-time deployment due to their lack of interpretability. Therefore, employing a complex CNN-LSTM model could be an efficient approach for implementing DL in small datasets. The utilization of AI in medical imaging shows great potential \cite{ramaekers2023computer}. 

The proposed research work offers two unique contributions:
\begin{itemize}
     \item It makes deep learning models for IVF embryo selection more transparent and reliable.
     \item It introduces a powerful LIME-based method that greatly improves the accuracy and clarity of interpreting complex embryo images.
\end{itemize}

The article is organized as follows: Section \ref{sec:2} provides a review of existing literature concerning embryo classification and XAI implementations. Section \ref{sec:3} outlines the methodology employed in this study. Section \ref{sec:4} presents the datasets and experimental results for embryo classification and LIME interpretation. Finally, Section \ref{sec:5} summarizes the research's main findings and discusses their implications for future research in the field.

\section{Related Work} \label{sec:2}
Recent advancements in AI indicate that in many cases DL approaches go beyond human visual capabilities and achieve better performance in terms of accuracy and consistency \cite{liu2019comparison}. Researchers have utilized the benefits of machine learning (ML) and DL in analyzing images of blastocysts, primarily aiming to predict the most suitable one for implantation  \cite{kallipolitis2022explainable}. Tran et al. \cite{tran2020swot} conducted a SWOT analysis on human and ML-based embryo assessments, distinguishing between conventional and artificial intelligence-assisted medicine. After thoroughly assessing each approach, this paper recommends strategic measures to progressively shift from conventional to ML-based approaches. A disadvantage of this study is that the SWOT analysis may not account for practical challenges in integrating ML-based systems into existing medical workflows, including issues related to data quality, implementation costs, and the need for significant training and adaptation among healthcare professionals. The study by Wang et al. in 2021 \cite{wang2021deep}, utilized VGG-16 with modifications, achieving up to 89.9\% accuracy and incorporating Grad-CAM for visual explanations. However, Grad-CAM's inability to accurately localize cells in complex embryo images limits its interpretability in IVF applications. Recent studies have highlighted the efficacy of deep learning architectures, in capturing both spatial and temporal features from medical images, even when dealing with relatively small datasets \cite{shamrat2025explainable, 10691647, sutradhar2024advancing, 10546976}.

Furthermore, Chai et al. \cite{chai2022optimizing} effectively leverage ML and DL models, including a CNN, to improve blastocyst quality assessment in IVF, achieving a test accuracy of 83.33\%. While the study showcases advancements in model optimization, it falls short in accuracy for clinical use and lacks model interpretability, which is critical for ensuring transparency and trust in AI-driven medical decisions. The study by Rahman et al. \cite{rahman2023explainable} proposed an automated system for grading blastocyst embryos in IVF using CNN and VGG-16 with new classification layers, achieving high accuracies of 90\% and 94\% in testing, respectively. However, the VGG-16 model's high computational costs, large size, and lack of interpretability may limit its practicality and transparency in clinical applications. Alkindy et al. \cite{alkindy2023automated} utilized two pre-trained CNN models, Xception and ResNet50, for classifying day-3 embryos, with Xception outperforming ResNet50. Similarly, Eswaran et al. \cite{eswaran2023assessment} proposed a robust algorithm using DenseNet 201, achieving a notable accuracy of 92\%. Both studies employed pre-trained models without incorporating interpretability techniques. The use of interpretative methods in embryo selection for IVF draws inspiration from Rahman et al. \cite{rahman2023explainable}, who utilized machine learning and deep learning to evaluate pregnancy risk factors. Their study, which identified gradient boosting with 90.64\% accuracy as the top-performing model, emphasized the importance of early diagnosis in mitigating maternal risks. They enhanced prediction interpretability with LIME and SHAP, paving the way for similar explainable AI approaches in IVF embryo selection to improve transparency and decision-making.\\

\vspace{1mm}
\textbf{Problem Statement:} The automated selection of embryos for IVF treatment is an underexplored area, with limited available datasets and a lack of transparency in existing models. As IVF serves a significant portion of the population, there is a critical need for more reliable and interpretable methods. Current interpretability techniques, such as Grad-CAM, fail to accurately localize cells in complex embryo images. This study addresses these challenges by proposing a LIME-based approach, which offers enhanced localization and improved interpretability for clinical application.

The following section will detail how the proposed model interpretation method effectively resolves these challenges and advances embryo selection processes.

\section{METHODOLOGY}  \label{sec:3}
\subsection{Dataset} \label{sec:3.1}
The dataset employed in this study is the STORK (segmentation of temporal ovarian regions with K-means) dataset, outlined in the GitHub repository \cite{STORK}. The dataset comprises images of embryos that have undergone IVF. 
\begin{figure}[h]
\centering
\includegraphics[width =0.45\textwidth]{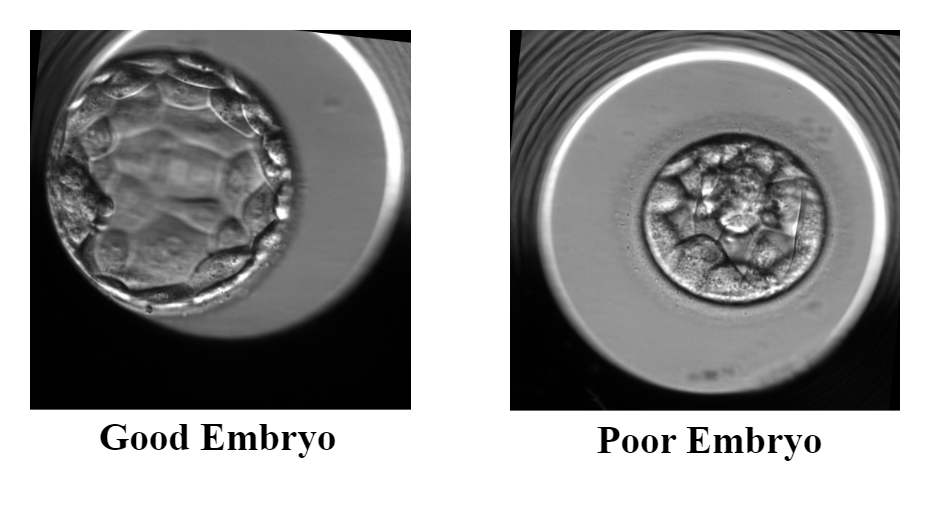}
\caption{Representation of good and poor embryos in STORK dataset}
\label{fig1:inout}
\end{figure} 

The dataset utilized in this study comprises 98 scan images of embryos, categorized into two classes: good embryo and poor embryo, each consisting of an equal number of 49 scan images. The images that are categorized as good embryos have been found to carry a higher likelihood of resulting in a live birth, while images that are identified as poor embryos are correlated with a lower probability of live birth. With that smaller number of samples, the model can’t be trained efficiently as the given dataset is highly constrained for model training. Since the available dataset here is insufficient for training deep neural networks (DNNs), additional synthetic data was generated using image augmentation techniques. Image augmentation proves as an advantageous technique to increase the samples in the dataset, thereby enhancing the effectiveness of model training.

\begin{table}[h!]
\centering
\caption{Description of STORK image dataset}
\label{table1}
\begin{tabular}{|c|cc|cc|c|}
\hline
\multirow{2}{*} {}  & \multicolumn{2}{c|}{\textbf{Good Embryos}} & \multicolumn{2}{c|}{\textbf{Poor Embryos}} & \multirow{2}{*}{\begin{tabular}[c]{@{}c@{}}\textbf{Total}\\ \textbf{Images}\end{tabular}} \\  \cline{2-5}
                             & \multicolumn{1}{c|}{Training}   & Testing  & \multicolumn{1}{c|}{Training}   & Testing  &                                 \\ \hline
\begin{tabular}[c]{@{}c@{}}\textbf{Before}\\\textbf{Augmentation} \end{tabular} & \multicolumn{1}{c|}{39}         & 10       & \multicolumn{1}{c|}{39}         & 10       & 98                              \\ \hline
\begin{tabular}[c]{@{}c@{}}\textbf{After}\\\textbf{Augmentation}\end{tabular}  & \multicolumn{1}{c|}{588}        & 147      & \multicolumn{1}{c|}{588}        & 147      & 1470                            \\ \hline
\end{tabular}
\end{table}

\begin{figure*}[!h]
\centering
\includegraphics[width =0.9\textwidth]{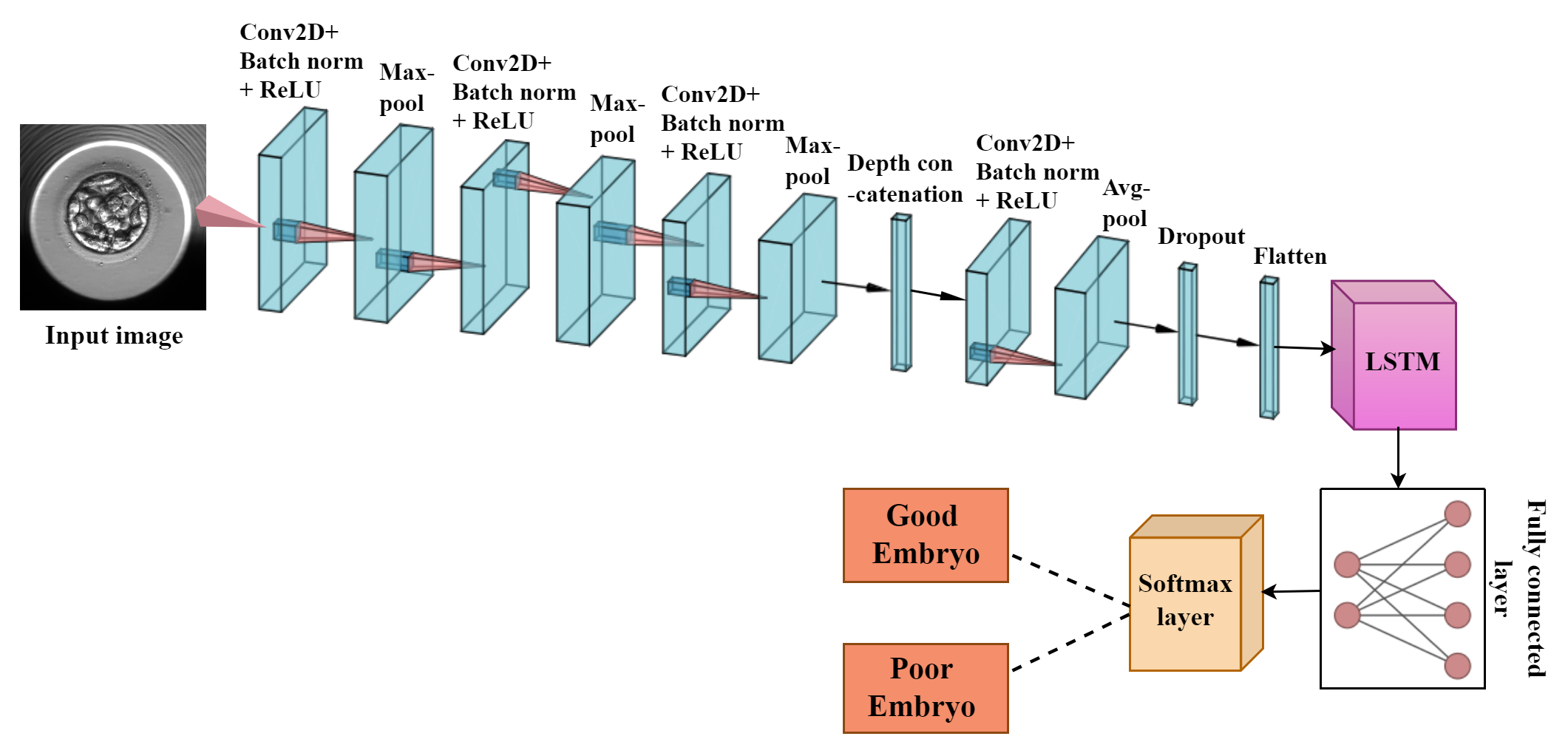}
\caption{Block diagram of the proposed CNN-LSTM model in selection of embryos}
\label{fig2:CNNLSTM}
\end{figure*}

\subsection{Image augmentation}
            Image augmentation refers to the process of expanding the training dataset through the use of diverse randomized alterations to the existing images. Table \ref{table1} describes the description of the data before and after augmentation. Geometric transformations can be used to enhance the diversity within the dataset \cite{xu2023comprehensive}. To increase the variation in the dataset, specific types of image augmentation techniques like rotation and reflection are applied, so that a diverse dataset of 1470 images was created. In the augmentation process for image datasets, random rotation within the range of -10 to 10 degrees is applied to the images with a probability of 0.5. Reflection, on the other hand, is applied with a probability of 1. This process introduces variations such that the dataset transforms, contributing to the creation of 1372 images randomly derived from the original images. This enhances the capacity of deep learning models to generalize effectively and enhance their performance, ultimately contributing to more precise IVF treatment. The synthetic data created through image augmentation can be utilized for the training of DNNs and can help to improve the performance in generalization and make efficient predictions on unseen data.

\subsection{CNN-LSTM Model}
A basic convolutional neural network architecture alone could not be able to recognize all crucial features and important details present in the images, leading to insufficient performance of the model. To achieve improved performance of the model and also increase the accuracy, we are using a fusion of  convolutional neural network (CNN) and long short-term memory (LSTM) architecture, known as CNN-LSTM, depicted in Fig. \ref{fig2:CNNLSTM}. The CNN-LSTM model processes the entire blastocyst image, allowing it to automatically learn and focus on important features across the image without the need for separate segmentation. This method is designed for the efficient classification of embryos from the features, which are acquired from the scanned images of embryos. This advanced model is strategically designed to excel in efficiently classifying embryos, leveraging intricate features extracted from images.

\begin{table*}[!h]
\centering
\caption{Description of layers and learnable parameters with CNN-LSTM model}
\label{table2}
\begin{tabular}{|c|c|c|}
\hline

Name & Layers & Learnable parameters \\ \hline
Image input layer & 2D image & - \\ \hline
Convolution layer & \begin{tabular}[c]{@{}c@{}}Conv2D layer (3x3x32) with stride {[}1 1{]}\\ Batch norm with 32 channels \\ ReLU\end{tabular} & \begin{tabular}[c]{@{}c@{}}Weights \& bias \\ Offset \& scale\\ -\end{tabular} \\ \hline
Convolution layer ( x 4) & \begin{tabular}[c]{@{}c@{}}Conv2D layer (3x3x64) with stride {[}1 1{]}\\ Batch norm with 64 channels \\ ReLU\\ \end{tabular} & 
\begin{tabular}[c]{@{}c@{}}Weights \& bias \\ Offset \& scale\\ -\end{tabular} \\ \hline
Convolution layer ( x 4) & \begin{tabular}[c]{@{}c@{}}Conv2D layer (3x3x32) with stride {[}1 1{]}\\ Batch norm with 32 channels \\ ReLU\\
Max-pool layer (3x3) with stride {[}2 2{]}\end{tabular} & 
\begin{tabular}[c]
{@{}c@{}}Weights \& bias \\ Offset \& scale\\ -\\ -\end{tabular} \\ \hline
Convolution layer ( x 2) & \begin{tabular}[c]{@{}c@{}}Conv2D layer (3x3x64) with stride {[}1 1{]}\\ Batch norm with 64 channels \\ ReLU\\
Max-pool layer (5x5) with stride {[}2 2{]}\end{tabular} & 
\begin{tabular}[c]
{@{}c@{}}Weights \& bias \\ Offset \& scale\\ -\\ -\end{tabular} \\ \hline
Convolution layer ( x 2) & \begin{tabular}[c]{@{}c@{}}Conv2D layer (3x3x64) with stride {[}1 1{]}\\ Batch norm with 32 channels \\ ReLU\\ Avg-pool layer (5x5) with stride {[}2 2{]}\end{tabular} & \begin{tabular}[c]{@{}c@{}}Weights \& bias \\ Offset \& scale\\ -\\ -\end{tabular} \\ \hline
Depth concatenation & - & - \\ \hline
Dropout & 40\% dropout layer & - \\ \hline
Flatten & Flatten layer & - \\ \hline
LSTM & LSTM with 128 hidden units & \begin{tabular}[c]{@{}c@{}} Input weights, \\recurrent weights,\\ \& bias \end{tabular} \\ \hline
Fully connected layer & 2 FC layer & Weights \& bias \\ \hline
Soft-max layer & - & - \\ \hline
Classification output & - & \begin{tabular}[c]{@{}c@{}} Cross-entropy \\with output labels \end{tabular}\\ \hline
\end{tabular}
\end{table*}

The proposed model comprises of 13 convolutional layers, with diverse configurations of 32 and 64 filters of size [3 3]
across distinct layers. The raw scanned images of embryos are
provided as the input for the CNN, subsequently implementing
an intermediate batch normalization technique after every
convolution layer to facilitate accelerated learning. Besides
that, an activation function, rectified linear unit (ReLU) is
employed after each convolution operation, thereby proposing
nonlinearity to the network. Max-pooling and average-pooling
techniques are also included to speed up the extraction of
features of embryo images. The max-pooling approach selects
maximum values within each window of size [3 3] and [5 5],
resulting in extracting prominent characteristics. Conversely, the average pooling method computes the average value
within each window of size [5 5], considering all values
with comparable importance. Depth concatenation operation is also included, which merges the output features. To mitigate
overfitting, a regularization technique, the dropout layer, is
employed. This layer introduces a type of redundancy in the
network to tackle the problem of model’s performance not
being prominent on unseen data. Then, the features obtained
from several convolutional layers are presented to flatten the layer, converting the data into linear form (1 dimensional)
while preserving the spatial information. Table \ref{table2} comprises of configuration details of the network.

LSTM, a variant of recurrent neural network (RNN), was specifically developed to overcome the issue of the vanishing gradient that may arise during
the training of conventional RNNs \cite{radha2024raw}. During training, neural networks employ optimization algorithms based on gradients to enhance their weights and reduce the impact of the loss function. The network’s weights are adjusted in the opposite direction of the gradient to reduce the loss function. In a standard RNN, the gradients of the loss function, calculated through backpropagation, can diminish significantly
as they propagate backward, resulting in an inability to capture long-term dependencies. To address this concern, LSTMs incorporate a gating mechanism that empowers them to decide which information to retrieve and what to discard from the
preceding time steps \cite{divakar2024explainable}. LSTMs are equipped with memory cells embedded within each recurrent unit. These memory cells serve the function of storing and preserving information over time, thereby enhancing the network’s capacity to capture and retain distinct dependencies. To regulate the flow of information within the network, LSTMs employ three gates: the input gate, the forget gate, and the output gate. Incoming information is supervised by the input gate, while outgoing information is overseen by the forget gate. The output gate controls the information retained within the cell over time. LSTM cells possess a cell state that functions as an internal memory for the cell. Through the utilization of gates, LSTM modifies its cell state, allowing it to either retain or discard information from preceding time steps.

The CNN-LSTM model employed in this study comprises approximately 15-20 million trainable parameters. This parameter count accounts for the convolutional layers, the LSTM layer with 128 hidden units, and the fully connected layers, all of which contribute to the model's ability to classify embryo images. Training was conducted on a Dell Precision 7960 Tower Workstation equipped with dual NVIDIA RTX A6000 GPUs, each with 48 GB of GDDR6 memory. The system's 256 GB of DDR5 RAM and high-speed NVMe SSD storage further facilitated efficient model training. A dropout rate of 0.4 was applied before the LSTM layer to prevent overfitting. The Adam optimizer was chosen due to its adaptive learning rate properties, with an initial learning rate of 0.001 and a piecewise learning rate schedule that reduces the rate every five epochs. The mini-batch size was set to 32, optimizing memory usage and convergence speed. These settings were fine-tuned through multiple iterations to achieve the best classification performance. The total training time for 25 epochs per fold with 10 folds and 18 seconds per epoch would be approximately 1.25 hours. The output from the LSTM layer is directed into 2 fully
connected layers (FC), thereby enhancing the model’s capability to comprehend complex data relationships, incorporate non-linearity, and address a variety of tasks. Subsequently, a softmax layer is incorporated for embryo classification.
It employs the softmax function on raw scores to produce probability vectors, to categorize input images into predefined classes. Each layer possesses learnable parameters that undergo fine-tuning throughout the training process, as shown in Table \ref{table2}.

\subsection{LIME}
Besides enhancing model performance, an intensifying concern in the application of DL in the field of medicine revolves around enhancing the model's trustworthiness. 
The black-box nature of DL models arises from their extensive parameters and complex structure. Even if a DL-based embryo classification model achieves high accuracy, it remains challenging for doctors or patients to trust its validity unless it can be explained. Hence, it is necessary to incorporate XAI technology into embryo classification to comprehend the evolving complex models. XAI involves a range of techniques/methods, intended to illuminate the decision-making process of a specific DL models. When the model is used for making decisions in real-time environments, the end users might be doubtful regarding the predictions of the model. So, it is important to comprehend the underlying reasons for the predictions that are made by the model, particularly in real-time environments such as the medical field. Understanding the reasons behind the model's predictions also yields insights into the model, which in turn converts the unreliable model into a reliable one. Therefore, providing explanations behind the predictions is essential to establish trust in the process among humans. 
In medicine, there is an increasing need for AI methodologies that not only demonstrate high performance but also offer trustworthiness, transparency, interpretability, and explainability for human experts \cite{holzinger2019causability}. The term “understanding” in the perspective of XAI refers to attaining a practical understanding of the model, which helps to describe the model’s black-box nature. Moreover, the use of LIME for interpretability in clinical image processing has been validated in various works, demonstrating its capability to offer human-understandable insights from black-box models \cite{10546976,10691647}.

\begin{figure*}[!h]
\centering
\includegraphics[width =0.9\textwidth]{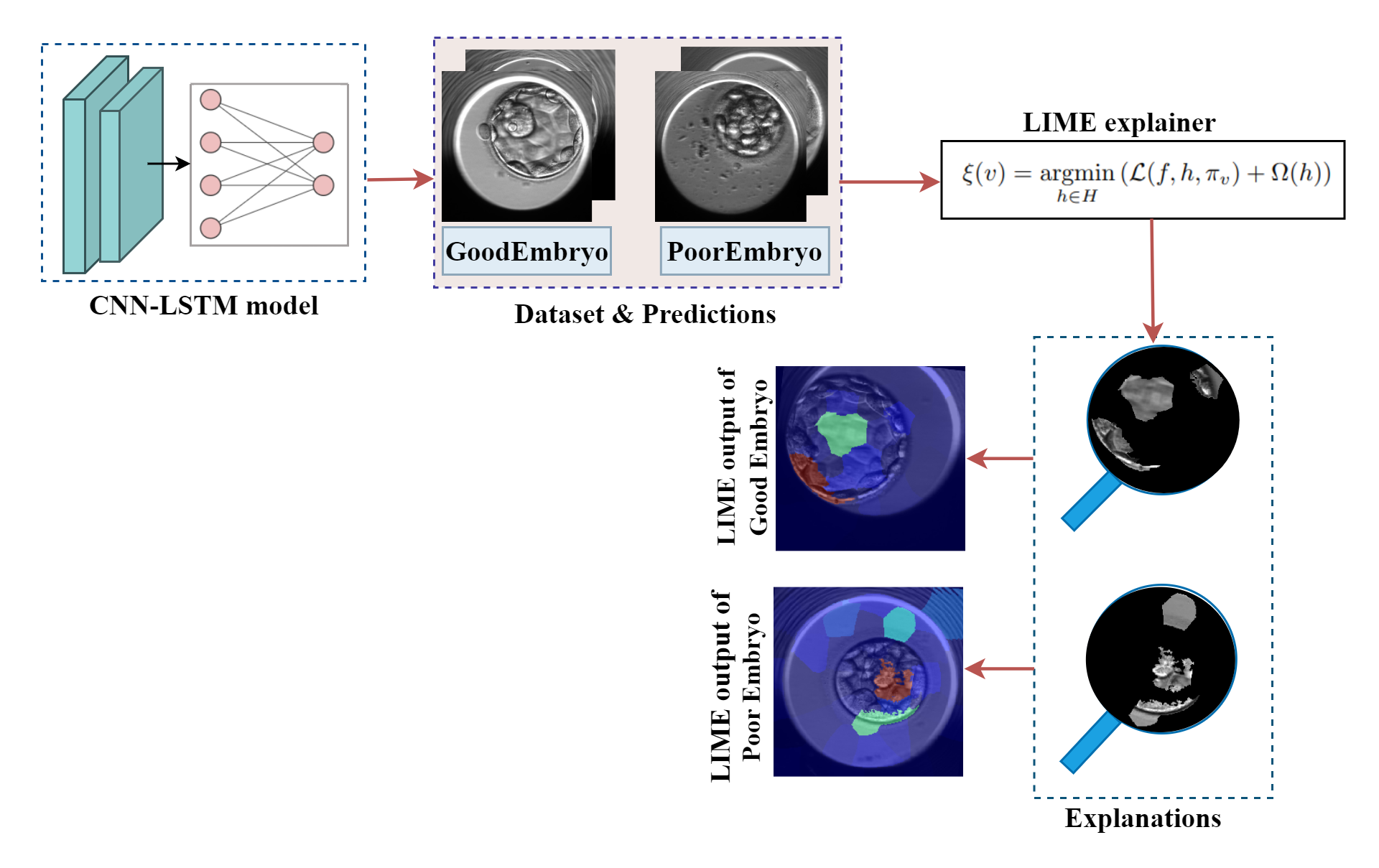}
\caption{Generation of visual explanations using LIME}
\label{fig2:lime}
\end{figure*} 

LIME, which stands for local interpretable model agnostic explanations, serves as a widely used tool to offer local explainability for the model. It not only offers user-friendly explanations of image data but also assists in understanding the black-box characteristics of DL models. This algorithm effectively interprets any classifier by deploying localized interpretable models with approximations, assuring faithful representations. It provides a global perspective to nurture confidence in any mysterious model, enabling the recognition of interpretable models described by local trustworthiness to the algorithm, instead of depending on human interpretable representations. Its primary operation involves attaining interpretable data representations, achieving an equilibrium between the truthfulness and interpretability of the model, and actively acquiring in-depth investigations at a local level. Lime efficiently distinguishes significant features as depicted in Fig. \ref{fig2:lime} and chooses interpretable data representations that are understandable to novice users. When describing the models that are trained on image data, the core algorithm might utilize complex feature vectors in its decision-making mechanism. However, those feature vectors might be difficult to understand for users lacking technical knowledge. Therefore, if explainability is conveyed through the existence or absence of region of interest, it constitutes a human-interpretable approach to explaining the model. 

Lime utilizes inherently interpretable models such as decision trees, linear models, and rule-based heuristic models to offer explanations to non-expert users, with visual artifacts \cite{bhattacharya2022applied}.  Assuming that the model being explained is denoted as $f: \mathbb{R}^d \rightarrow \mathbb{R}$. In image classification, an interpretable representation of the complex model is conveyed through a binary vector, which is expressed as $ v' \in \{0,1\}^d $. The initial representation of the image data instance targeted for explanation is denoted as $v \in \mathbb{R}^d$, where $d$ signifies the dimension of the data. This explanation represents a model, that is specified by $ h \in H$, where $H$ illustrates the class of interpretable models. The domain of $h$ is depicted by another binary vector, $\{0,1\}^{d'}$, indicating the presence or absence of comprehensible components. Besides the interpretability, this algorithm tries to measure the complexity of the explanation, expressed mathematically as 
$\Omega(h)$. 
Lime preserves local fidelity while delivering explanations \cite{bhattacharya2022applied}. Therefore, to measure the closeness among data instances, a mathematical function expressed as $\pi_v(z)$ is used, which further symbolizes the closeness surrounding the original representation. Given a probability function, $f(v)$, indicating the likelihood of $v$ belonging to a specific class. To estimate $f$, LIME aims to calculate the level of unfaithfulness in $h$ using a proximity function, $\pi_v$. This complete operation is symbolized by the function $\mathcal{L}(f,h,\pi_v)$. To obtain an equilibrium between local fidelity and interpretability, we need to minimize $\mathcal{L}(f,h,\pi_v)$ while maintaining $\Omega(h)$ as low as possible for human comprehension. The explanation generated by LIME is acquired as follows:
\begin{equation}
    \xi(v) = \underset{h \in H}{\text{argmin}} \left( \mathcal{L}(f, h, \pi_v) + \Omega(h) \right) \label{equ1}
\end{equation}
Therefore, from the above equation, it can be asserted that trade-off measure relies on the interpretable models $(H)$, the fidelity function $(\mathcal{L})$, and the complexity measure $(\Omega(h))$.

\begin{table*}[!h]
\centering
\caption{Performance analysis of the  CNN-LSTM model in terms of percentage (\%)}
\label{table3}
\begin{tabular}{|c|c|c|c|c|c|}
\hline
                    & \textbf{Accuracy} & \textbf{Precision} & \textbf{Recall} & \textbf{F1-score} & \textbf{Sensitivity} \\ \hline
Before augmentation & 90             & 83.3              & 100               & 90.9             & 100                   \\ \hline
After augmentation  & 97.7            & 100                 & 95.4           & 97.6             & 95.4                \\ \hline
\end{tabular}
\end{table*}

 \begin{figure*}[!h]
\centering
\includegraphics[width=0.89\textwidth]{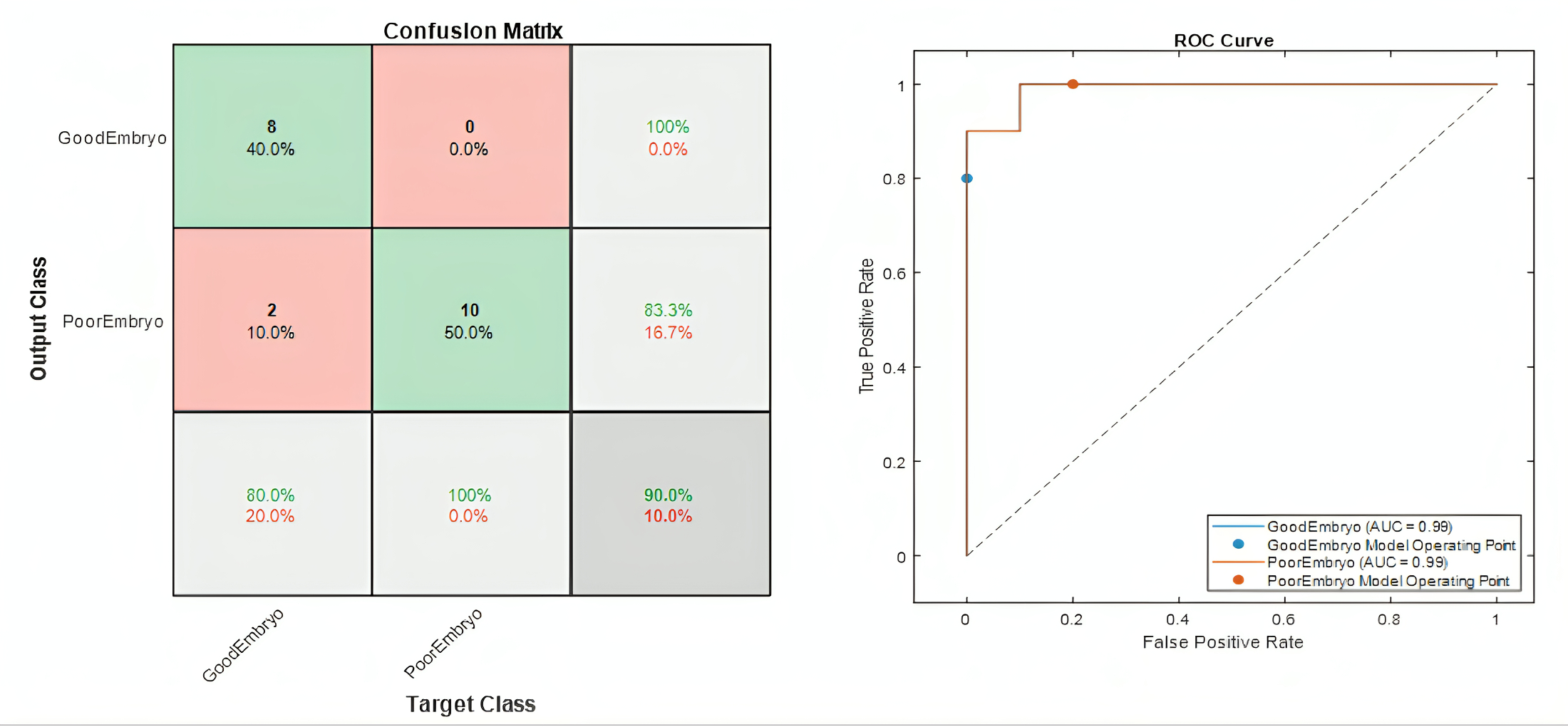}
\caption{Performance analysis of CNN-LSTM model before augmentation in terms of (a) confusion chart (b) ROC curve}
\label{fig2:plot}
\end{figure*}

\begin{figure*}[!h]
\centering
\includegraphics[width=0.89\textwidth]{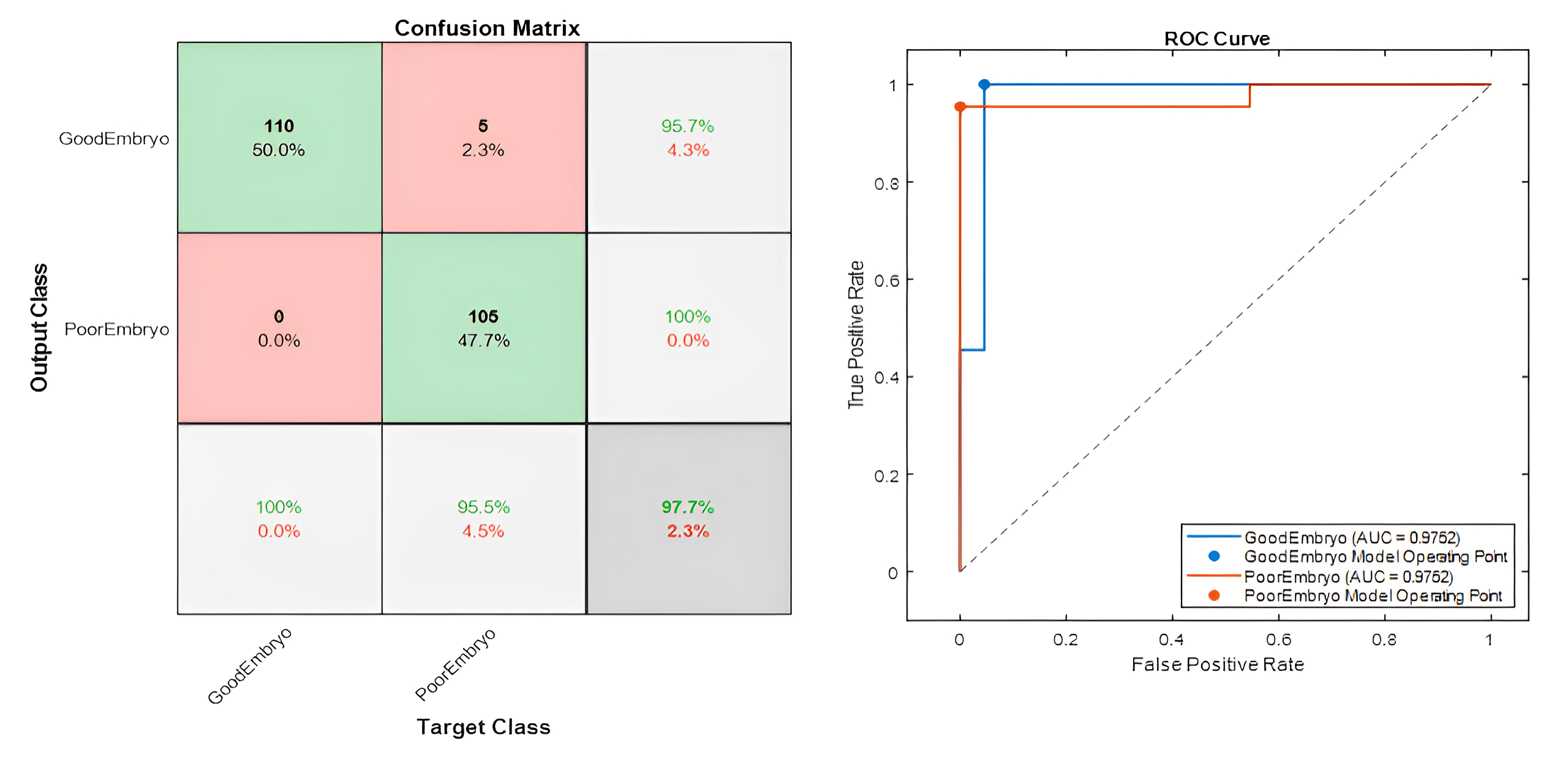}
\caption{Performance analysis of CNN-LSTM model after augmentation in terms of (a) confusion chart (b) ROC curve}
\label{fig3:plot}
\end{figure*} 

\begin{figure*}[h]
    \centering
{\includegraphics[width=\textwidth]{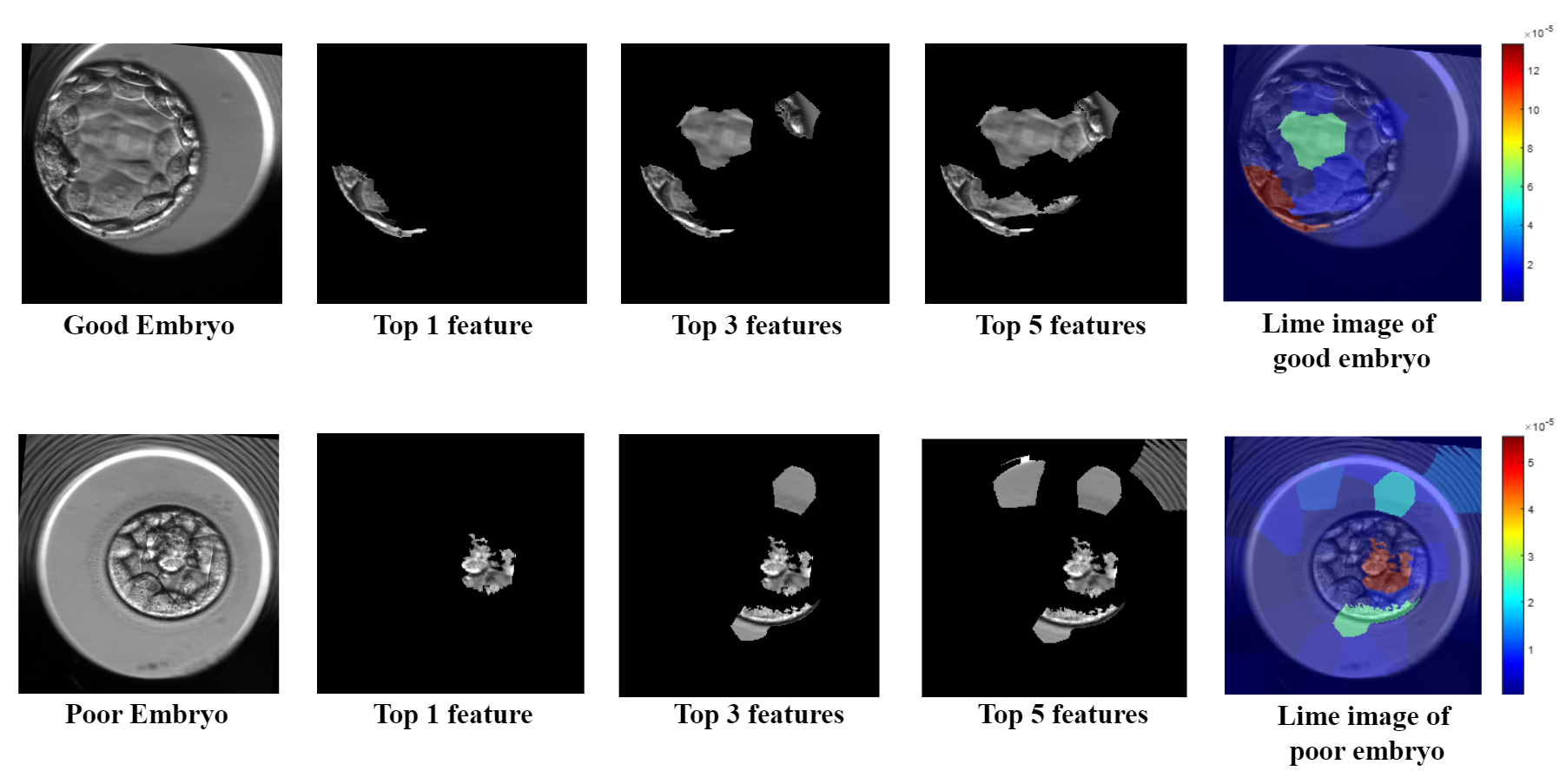}}
    \caption{The top row represents the good embryo class, accompanied by top most features and visual explanations and the next row represents the poor class with top important features and visual explanations}
    \label{fig:fig_2}
\end{figure*}

\begin{figure*}[!h]
\centering
{\includegraphics[width=0.9\textwidth]{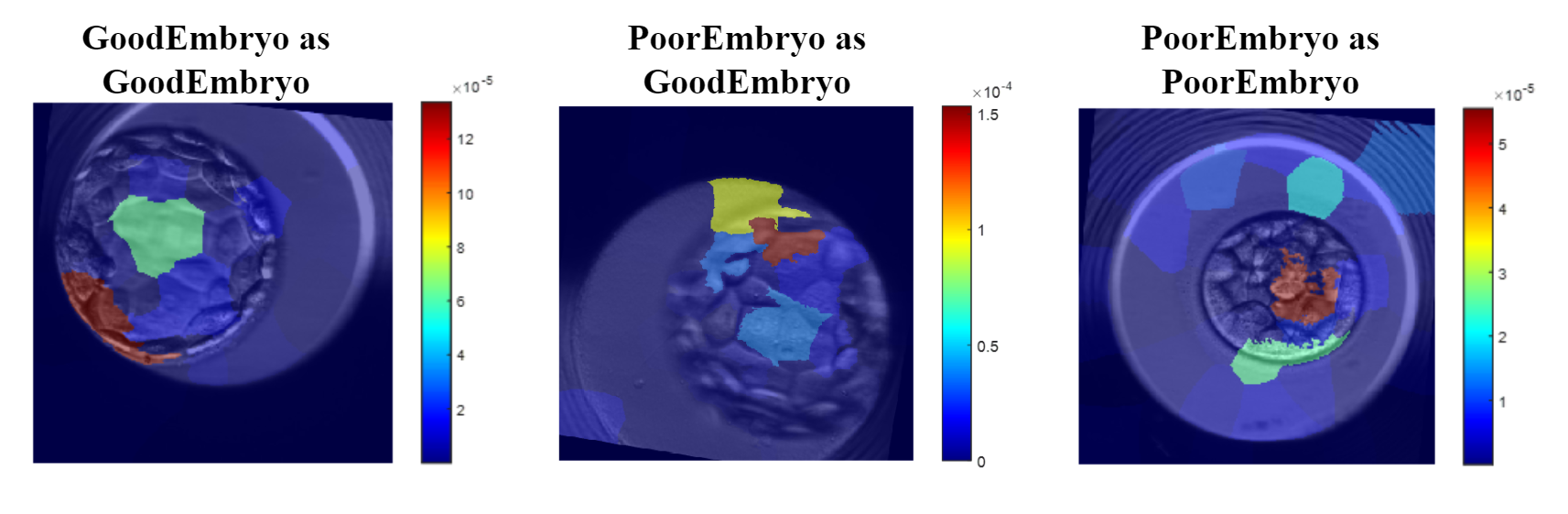}}
    \caption{LIME visualizations—Correct classifications on left and right, misclassification in center}
    \label{fig:fig_miss}
\end{figure*}

The proposed complex model is represented by
$f$, while the simple model (local model) is represented by $h$, which is selected from a set of interpretable models denoted by $H$. The family $H$ comprises linear models, such as linear regression, and its various variants.
 $\pi_v$ defines the local neighborhoods of the data points, determining their proximity within the data points. The term $\Omega(h)$ regulates the complexity of the simple local model, particularly in the context of linear regression, serving as a metric around which the  optimization problem revolves to minimize. The objective is to encourage the inclusion of numerous zero-weighted input features, thus reducing the overall complexity of the model. The above equation expresses a simple model ($h$), that minimizes the loss term, seeking to estimate the complex model in that specific region while maintaining simplicity to the greatest extent possible. To compute the initial loss term, LIME initially creates new data points in the proximity of the image input data. More specifically, LIME randomly generates data points across the dataset, but their weights are determined by the distance to our input data, emphasizing the local regions surrounding the input data point, striving to understand the small details and differences nearby. These data points are generated through perturbations, achieved by sampling from a normal distribution characterized by the mean and standard deviation for each feature. Let's consider a perturbed sample containing fractions of non-zero elements of $z'$ denoted by $ z' \in \{0,1\}^d $.  The algorithm strives to recover samples from the original representation, to approximate $f(z)$. Subsequently, predictions are obtained for these data points using our complex model $f$. As the new data points are generated through perturbations, using this as a new dataset, we can train a classifier, subsequently, $f(v')$ is employed as a label for the explanation model, $\xi(v)$, that are derived from predictions made by the complex model $f$. Minimizing the first loss term involves achieving the highest accuracy on the new dataset by employing a simple linear model. 

 H represents the class of linear models, such that $h(z') = w_g \cdot z_0$. We adopt the locally weighted square loss as $\mathcal{L}$, as defined in Eq. \ref{equ2} and define $\pi_v(z) = \exp\left(-\frac{D(v, z)^2}{\sigma^2}\right)$ representing an exponential kernel concerning a distance function D with width $\sigma$. 
 \begin{equation}
     \mathcal{L}(f, h, \pi_v) = \underset{z, z_0 \in Z} \sum\pi_v(z) (f(z) - h(z'))^2 \label{equ2}
 \end{equation}

 For image classification, we utilize $\Omega$ for superpixels, establishing a framework where each super-pixel acts as a fundamental unit for analysis. The interpretable representation of an image is depicted as a binary vector, where 1 corresponds to the original super-pixel and 0 denotes a grayed-out super-pixel. The specific selection of $\Omega$ makes it challenging to directly solve Eq. \ref{equ1}. However, we opt to approximate it by first choosing K features using lasso regression, subsequently learning the weights through least squares.

\section{Results \& Discussion}  \label{sec:4}
In this section, we present an overview and analysis of the results derived from our investigations into the model’s performance both before and after applying image augmentation. Before augmentation, the dataset comprised 98 embryo images, 49 images per class, ensuring a balanced representation of both categories: good and poor embryos.

\begin{table*}[!b]
\small
\centering
\caption{Comparison of existing literature and their performance in embryo classification}
\label{table4}
\begin{tabular}{|c|c|c|c|c|c|}
\hline
\multicolumn{1}{|c|}{\begin{tabular}[c]{@{}c@{}}\textbf{Author/Reference}/\\ \textbf{Year}\end{tabular}} & \multicolumn{1}{c|}{\textbf{Dataset}}                                                                                                    & \multicolumn{1}{c|}{\textbf{Augmentation}}                                                & \multicolumn{1}{l|}{\textbf{Model}}                                                                                                                                                                               & \multicolumn{1}{l|}{\textbf{Accuracy}}                                                                             & \multicolumn{1}{c|}{\textbf{\begin{tabular}[c]{@{}c@{}}Presence \\ of XAI\end{tabular}}} \\ \hline
\begin{tabular}[c]{@{}c@{}} Thirumalaraju et al.\cite{thirumalaraju2021evaluation}\\ 2021\end{tabular}             & \begin{tabular}[c]{@{}c@{}}3,469 recorded videos \\ collected from 543 \\ patients from general \\ hospital fertility \\ center, Massachusetts\end{tabular}                                               & No                                                                                        & \begin{tabular}[c]{@{}c@{}}Inception-v3,\\ ResNET-50,\\ Inception-ResNET-v2,\\ NASAnetLarge,\\ ResNeXt-101,\\ ResNeXt-50,\\ Xception\end{tabular}                                                                & \begin{tabular}[c]{@{}c@{}}89.08\%,\\ 89.71\%,\\ 90.12\%,\\ 78.44\%,\\ 90.75\%,\\ 89.95\%,\\ 90.48\%\end{tabular} & Saliency maps                                                                            \\ \hline
\begin{tabular}[c]{@{}c@{}} Vaidya et al.\cite{vaidya2021time}\\ 2021\end{tabular}                     & \begin{tabular}[c]{@{}c@{}}5458 images collected \\ from nova IVF \\ fertility, Ahmedabad\end{tabular}                                                                                                    & No                                                                                        & \begin{tabular}[c]{@{}c@{}}VGG 16-LSTM,\\ VGG 19-LSTM,\\ Xception-LSTM,\\ Inception-LSTM,\\ MobileNetV2-LSTM,\\ DenseNet21-LSTM,\\ VGG 16-GRU\end{tabular}                                                       & \begin{tabular}[c]{@{}c@{}}100\%,\\ 100\%,\\ 99.7\%,\\ 98.8\%,\\ 100\%,\\ 99.9\%,\\ 100\%\end{tabular}             & No                                                                                       \\ \hline
\begin{tabular}[c]{@{}c@{}}Wang et al.\cite{wang2021deep}\\ 2021\end{tabular}                      & \begin{tabular}[c]{@{}c@{}}11,275 images collected\\ from drum tower\\ hospital of Nanjing\\ university medical \\ school, China\end{tabular}                                                             & No                                                                                        & \begin{tabular}[c]{@{}c@{}}VGG 16-MobileNetV2,\\ VGG-16 \\ (voting machine),\\ VGG 16 (multiple\\ channel combination)\end{tabular}                                                                               & \begin{tabular}[c]{@{}c@{}}86.5\%,\\ 87.2\%,\\ \\ 89.9\%\end{tabular}                                              & Grad-CAM                                                                                 \\ \hline
\begin{tabular}[c]{@{}c@{}} Chai et al. \cite{chai2022optimizing}\\ 2022\end{tabular}                      & \begin{tabular}[c]{@{}c@{}}Publicly available dataset\\ consisting of 249 images\end{tabular}                                                                                                              & \begin{tabular}[c]{@{}c@{}}Sheared, zoomed\\ \& flipped\end{tabular}                      & \begin{tabular}[c]{@{}c@{}}CNN, SVM, \\ KNN, LR \& \\ GB \end{tabular}                                                                                                                              & \begin{tabular}[c]{@{}c@{}}84\%, 82\%,\\ 74\%, 82\%\\ \& 64\%\end{tabular}                                         & No                                                                                       \\ \hline
\begin{tabular}[c]{@{}c@{}} Paya et al.\cite{paya2022automatic}\\ 2022\end{tabular}                       & \begin{tabular}[c]{@{}c@{}}Private database was \\ created from IVF valencia, \\ consists of 3014 embryo \\ cycles for development \& \\ evaluation, 830 embryo \\ cycles for quality models\end{tabular} & \begin{tabular}[c]{@{}c@{}}Random flipping,\\ rotation\end{tabular}                       & \begin{tabular}[c]{@{}c@{}}Supervised contrastive\\ learning (to predict\\ embryo quality at day-4\\ and day-5)\\ Inductive transfer\\ learning (to classify \\ embryo quality at day-4\\ and day-5)\end{tabular} & \begin{tabular}[c]{@{}c@{}}81.03\% \& \\ 93.3\%\\ \\ \\ 75\% \& \\ 80.01\%\end{tabular}                            & Grad-CAM                                                                                 \\ \hline
\begin{tabular}[c]{@{}c@{}} Eswaran et al.\cite{eswaran2023assessment} \\ 2023\end{tabular}                   & \begin{tabular}[c]{@{}c@{}}4000 images collected\\ from publicly available\\ dataset\end{tabular}                                                                                                          & \begin{tabular}[c]{@{}c@{}}Rescale, rotate,\\ height shift \& \\ width shift\end{tabular} & \begin{tabular}[c]{@{}c@{}}DenseNet-201,\\ ResNet-50\end{tabular}                                                                                                                                                 & \begin{tabular}[c]{@{}c@{}}92.03\%,\\ 76.15\%\end{tabular}                                                         & No                                                                                       \\ \hline
\begin{tabular}[c]{@{}c@{}} Alkindy et al. \cite{alkindy2023automated}\\ 2023\end{tabular}                   & \begin{tabular}[c]{@{}c@{}}152 images of day-3\\ embryos collected from\\ Vistana fertility centre,\\ Malaysia\end{tabular}                                                                               & \begin{tabular}[c]{@{}c@{}}Flip, rotation,\\ zoom, shear, \& \\ filling mode\end{tabular} & \begin{tabular}[c]{@{}c@{}}ResNet50,\\ Xception\end{tabular}                                                                                                                                                      & \begin{tabular}[c]{@{}c@{}}75\%,\\ 98\%\end{tabular}                                                               & No                                                                                       \\ \hline
\begin{tabular}[c]{@{}c@{}} Mohamed et al.\cite{mohamed2023automated}\\ 2023\end{tabular}                   & \begin{tabular}[c]{@{}c@{}}110 images of day-5 \\ embryos collected from \\ Vistana fertility center,\\ Malaysia\end{tabular}                                                                              & \begin{tabular}[c]{@{}c@{}}Rotation, zoom,\\ shear, width shift\end{tabular}              & \begin{tabular}[c]{@{}c@{}}CNN,\\ VGG-16\end{tabular}                                                                                                                                                             & \begin{tabular}[c]{@{}c@{}}90\%,\\ 94\%\end{tabular}                                                               & No                                                                                       \\ \hline
Proposed Model                                                                         & \begin{tabular}[c]{@{}c@{}}98 original images available from \\ GitHub repository. \\ After augmentation 1470 images \\used for modeling\end{tabular}                                                                                                                                 & \begin{tabular}[c]{@{}c@{}} Rotation \& \\ reflection\end{tabular}                         & CNN-LSTM                                                                                                                                                                                                          & 97.7\%                                                                                                             & LIME                                                                                     \\ \hline
\end{tabular}

\end{table*}

During the training phase, the model was presented with a dataset of 78 images, evenly distributed with 39 images per class. Moreover, during the testing phase, 20 images from each class were utilized to evaluate the model’s performance, achieving an accuracy of 90\%. Following augmentation, the dataset expanded to incorporate 1470 images. During the training phase, the model was exposed to a dataset comprising 1176 images—588 images per class—contributing a well-balanced representation of two classes. Given the limited size of the original dataset, data augmentation was applied to introduce variability and improve the model’s generalization. However, synthetic augmentation cannot fully capture the complexity and diversity of real-world clinical data, which may limit the model's robustness in broader applications. To address this, 10-fold cross-validation was used on the training set to ensure rigorous performance assessment. This approach enabled the model to be trained and validated across multiple stratified subsets, reducing the risk of overfitting and improving generalizability. Subsequently, during the testing phase, the proposed CNN-LSTM model demonstrated a remarkable accuracy of 97.7\% when evaluated on a test set of 294 images. The performance evaluation of the proposed model before and after the augmentation technique is depicted through various evaluation metrics including accuracy, receiver operating characteristic (ROC) curve, precision, recall, sensitivity, and F1-score. Subsequently, during the testing phase, the proposed CNN-LSTM model demonstrated a remarkable mean accuracy of 97.7\% $\pm$ 0.82\% when evaluated on a test set of 294 images. This was determined based on five independent runs using different random initializations and data shuffling to ensure the robustness and consistency of the model’s performance. The performance evaluation of the proposed model before and after the augmentation technique is depicted through various evaluation metrics, including accuracy, receiver operating characteristic (ROC) curve, precision, recall, sensitivity, and F1-score.

 The confusion matrix depicted in Fig. \ref{fig2:plot} provides a detailed breakdown of the classification performance of the proposed model before augmentation. Individual class accuracies are as follows: good embryo achieved a flawless 80\% accuracy with two misclassifications as poor embryos, while poor embryos attained  100\% accuracy with no misclassifications. The confusion matrix presented in Fig. \ref{fig3:plot} indicates a detailed breakdown of the classification performance of the proposed model after the application of augmentation technique. The accuracies of individual classes are as follows: good embryo with 100\%, with no misclassification; poor embryo with 95.5\%, with five misclassifications as good embryo.  Precision, recall, F1-score, and sensitivity as shown in Table \ref{table3} stand as essential metrics for evaluating classification models, emphasizing their significance in model assessment. From Table \ref{table3}, it is evident that the model's performance metrics have improved after augmentation compared to the previous (before augmentation) results. To fulfill the need for interpretability, a novel, and agnostic, application of a specific XAI technique, LIME has been implemented. The provision of explanations will enhance the acceptance of the predictive model. Lime strives to mimic the model’s behavior in closeness to the individual data instances undergoing prediction. It essentially fits local interpretable models, especially within that local region. It aims to generate a local approximation of our complex CNN-LSTM model for a particular input for instance.  

By generating the local interpretable models, LIME highlights the important features that are crucial for the recognition of the classes. It spotlights the most significant attributes that classify the images as shown in Fig. \ref{fig:fig_2} The misclassifications were further analyzed using LIME, an XAI technique that produces visual explanations, to understand the specific regions in the input images that influenced the classification. Fig. \ref{fig:fig_2} illustrates the top 1, 3, and 5 important features crucial for predicting a specific class. The visual representations depict the crucial regions that are relevant for the classification of the embryo images. To quantitatively validate these explanations, a fidelity score of 0.91 was achieved, indicating that the local surrogate models accurately approximated the CNN-LSTM decision boundaries in most cases. Furthermore, the intersection-over-union (IoU) between the LIME-generated explanations and expert-annotated embryo regions yielded a mean IoU of 0.74, reflecting strong overlap between the model's focus areas and medically relevant regions identified by clinicians. These metrics substantiate the reliability and relevance of the LIME visualizations in clinical contexts. Specifically, LIME images representing the misclassifications between good and poor embryos exhibit patterns resembling those observed in the corresponding visual explanations.  This dual visual-quantitative interpretability approach provides valuable insights into the model’s attention mechanisms and supports the evaluation of both classification accuracy and decision-making transparency. Following augmentation, the model achieved 100\% accuracy for the good embryo class and 95.5\% accuracy for the poor embryo class, with five poor embryos misclassified as good as visualized by LIME in Fig. \ref{fig:fig_miss}. Clinically, such misclassifications can lead to the selection of non-viable embryos, reducing the chances of implantation success and increasing patient burden. Hence, reducing false positives is essential for ensuring the model’s reliability in real-world IVF applications.

\subsection{Comparative study}
In this section, a comparative evaluation of the proposed model against previous studies focusing on the grading of embryo classes using blastocyst images. 

The dataset specified within the GitHub repository \cite{STORK}, was utilized for training and testing our model. The proposed approach integrates the classification accuracy of the CNN-LSTM network with the interpretability facilitated by one of the XAI techniques, LIME. A comparison between the previously proposed models and the proposed model is depicted in Table \ref{table4}. It is important to note that model complexity metrics (e.g., parameter count or FLOPs) were not available for the baseline methods, as they were not reported in the original studies.

\section{CONCLUSION}  \label{sec:5}
This study highlights the importance of accurate and timely assessment in addressing challenges encountered in IVF treatment, which represents a significant global health concern. It illuminates the development of an automated system for evaluating the quality of embryos, aiming to improve the efficiency and accuracy of the birth rates in ART. Utilizing embryo images as an input, the CNN-LSTM model proposed in this study was trained on the augmented images derived from the raw embryo images for embryo grading. Interpretation of the proposed CNN-LSTM network is accomplished by using LIME, one of the most common XAI techniques. With the help of LIME, the decision-making process and the region of interests for classification can be visualized transparently, potentially enhancing its appropriateness for real-time clinical applications. LIME allows for visualizing the model’s decision-making procedure, enabling the provision of interpretable explanations customized for each instance. This integrated approach holds the potential for advancing embryo classification, enabling more informed decisions in medical interventions, and ultimately contributing to better outcomes for patients undergoing assisted reproductive procedures. Further, the proposed model can be extended by using larger datasets and diverse XAI techniques to improve interpretability. This will enable multi-class classification of blastocoel expansion stages, including early, mid, late, expanded, hatching, and hatched blastocysts.

\vspace{3mm}
\small{\textbf{Data Availability:} The open-access STORK dataset, which support the findings of this study, was sourced from the GitHub repository. Additionally, the full code is available on GitHub at \url{https://github.com/RadhaKodali/Explainable-AI-Models}}
\vspace{3mm}\\
\small{\textbf{Funding:} This research received no external~funding.} 
\vspace{3mm}\\
\small{\textbf{Declaration of Interests:}  The authors declare no conflict of interest.}

\bibliographystyle{unsrt}  
\bibliography{references}

\end{document}